\documentclass{article}

\usepackage[final]{neurips_2025}

\usepackage[utf8]{inputenc} 
\usepackage[T1]{fontenc}    
\usepackage[table]{xcolor} 
 \usepackage{float}

\usepackage{hyperref}       

\hypersetup{
    colorlinks=true,
    linkcolor=red,
    urlcolor=cyan,
    linktoc=all,
    citecolor=cyan
}

\usepackage{url}            
\usepackage{booktabs}       
\usepackage{booktabs}       
\usepackage{algorithm}
\usepackage{algpseudocode}
\usepackage{amsfonts}       
\usepackage{nicefrac}       
\usepackage{microtype}      

\usepackage{amsmath}         
\usepackage{bm}              
\usepackage{amssymb}
\usepackage{tabularx} 
\usepackage{multirow}
\usepackage{times}
\usepackage{epsfig}
\usepackage{graphicx}
\usepackage{adjustbox}
\usepackage{tabularray}
\usepackage{xspace}
\usepackage{caption}
\usepackage{stackengine}
\usepackage{makecell}
\usepackage{array}
\usepackage{utfsym}
\usepackage{bbding}
\usepackage{subfig}
\usepackage{enumerate}
\usepackage{wrapfig}
\usepackage{float} 
\usepackage{mathtools}
\usepackage{wrapfig}   
\usepackage{caption}   
\usepackage{enumitem}
\usepackage{cleveref}      

\usepackage{listings}
\definecolor{codegreen}{rgb}{0,0.6,0}
\definecolor{codepurple}{rgb}{0.58,0,0.82}
\definecolor{backcolour}{rgb}{0.95,0.95,0.92}

\definecolor{tabfirst}{rgb}{1, 0.7, 0.7}
\definecolor{tabsecond}{rgb}{1, 0.85, 0.7}
\definecolor{tabthird}{rgb}{1, 1, 0.7}

\makeatletter
\newcommand{\ie}{%
  \emph{i.e.}\@ifnextchar.{\@gobble}{}%
}
\newcommand{\eg}{%
  \emph{e.g.}\@ifnextchar.{\@gobble}{}%
}
\newcommand{\etc}{%
  etc\@ifnextchar.{}{\@.}%
}

\definecolor{myRed}{rgb}{0.8, .2, .2}
\definecolor{myGreen}{rgb}{0, .8, .3}
\definecolor{myOrange}{rgb}{0.7, 0.45, 0.2}

\newcommand{\methodName}{Hunyuan3D 2.1}

\title{Hunyuan3D 2.1: From Images to High-Fidelity 3D Assets with Production-Ready PBR Material}

\author{
Tencent Hunyuan \\
\\
\url{https://github.com/Tencent-Hunyuan/Hunyuan3D-2.1}
}

\begin{document}

\maketitle

\begin{figure}[h]
\centering
\includegraphics[width=\linewidth]{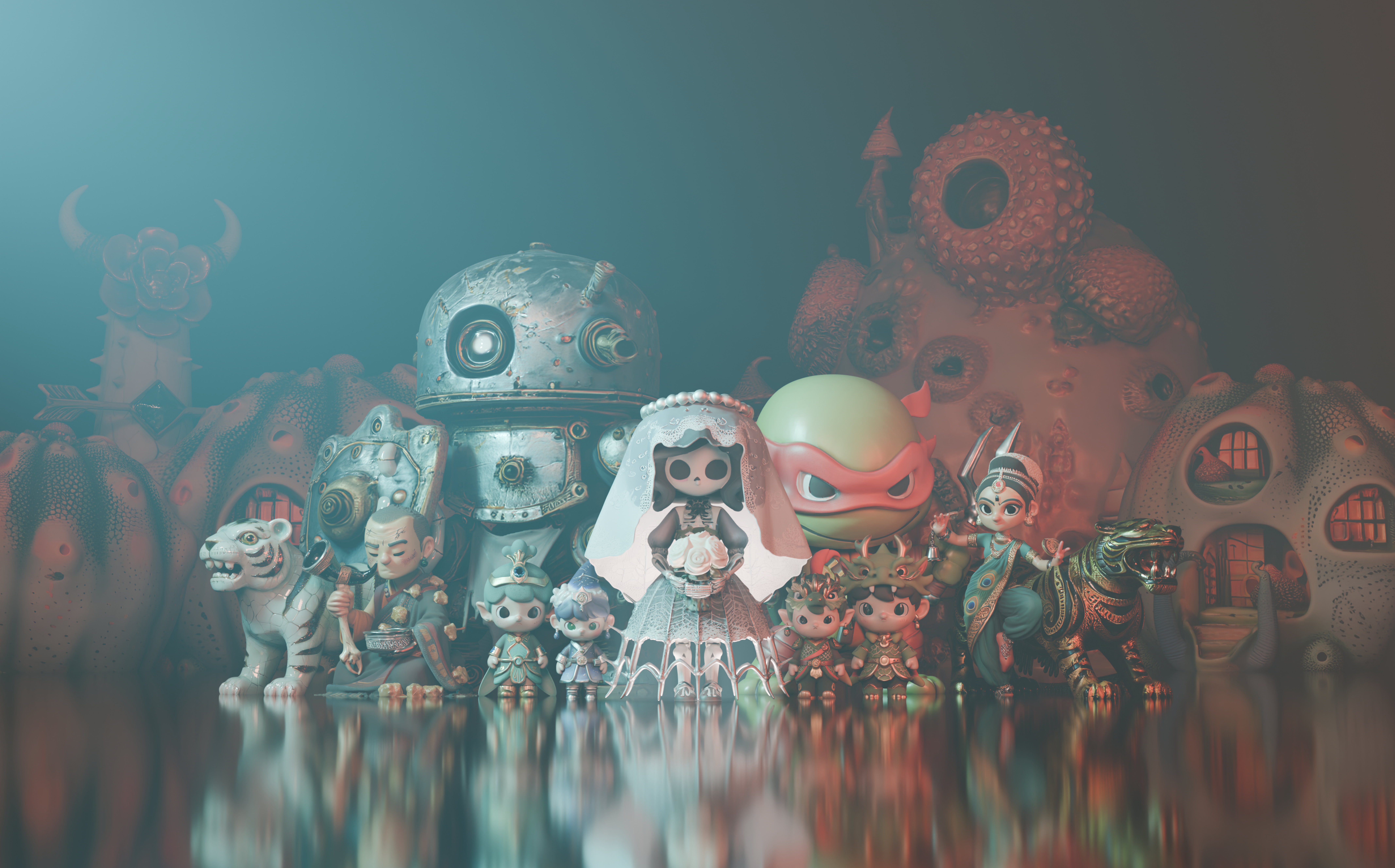}
\caption{Gallery of 3D assets generated by ~\textbf{\methodName{}.}}
\label{fig:teaser-top}
\end{figure}
 

\begin{abstract}
3D AI-generated content (AIGC) is a passionate field that has significantly accelerated the creation of 3D models in gaming, film, and design. Despite the development of several groundbreaking models that have revolutionized 3D generation, the field remains largely accessible only to researchers, developers, and designers due to the complexities involved in collecting, processing, and training 3D models. To address these challenges, we introduce ~\methodName{} as a case study in this tutorial. This tutorial offers a comprehensive, step-by-step guide on processing 3D data, training a 3D generative model, and evaluating its performance using ~\methodName{}, an advanced system for producing high-resolution, textured 3D assets. The system comprises two core components: the Hunyuan3D-DiT for shape generation and the Hunyuan3D-Paint for texture synthesis. We will explore the entire workflow, including data preparation, model architecture, training strategies, evaluation metrics, and deployment. By the conclusion of this tutorial, you will have the knowledge to finetune or develop a robust 3D generative model suitable for applications in gaming, virtual reality, and industrial design.
\end{abstract}

\section{Introduction}

While recent breakthroughs in 2D image and video generation—powered by diffusion models~\cite{ho2020denoising,rombach2022high,esser2024scaling,li2024hunyuandit,kong2024hunyuanvideo,wan2025}—have revolutionized content creation, the field of 3D generative modeling lags behind. Current methods for 3D asset synthesis remain fragmented, with incremental progress in foundational techniques such as latent representation learning~\cite{zhang20233dshape2vecset}, geometric refinement~\cite{zhao2023michelangelo,li2025triposg,weng2024scaling}, and texture synthesis~\cite{zhang2024clay,jiang2024flexitex,liu2024text}. Among these, CLAY~\cite{zhang2024clay} marks a milestone as the first framework to demonstrate the viability of diffusion models for high-quality 3D generation. Yet, unlike the thriving open-source ecosystems in image (~\eg, Stable Diffusion~\cite{rombach2022high}), language (~\eg, LLaMA~\cite{touvron2023llama}), and video (~\eg, HunyuanVideo~\cite{kong2024hunyuanvideo}, and Wan 2.1~\cite{wan2025}), the 3D domain lacks a robust, scalable foundation model to drive widespread innovation.

To bridge this gap, we introduce \emph{\methodName}, a comprehensive 3D asset creation system to generate a textured mesh from a single image input. It is mainly built on two fully open-source foundation models: 1) Hunyuan3D-DiT: A shape-generation model combining a flow-based diffusion architecture with a high-fidelity mesh autoencoder (Hunyuan3D-ShapeVAE); 2) Hunyuan3D-Paint: a mesh-conditioned multi-view diffusion model for PBR material generation, producing high-quality, multi-channel-aligned, and view-consistent textures.

For shape generation, we leverage Hunyuan3D-ShapeVAE and Hunyuan3D-DiT to achieve high-quality and high-fidelity shape generation. Specifically, Hunyuan3D-ShapeVAE employs mesh surface importance sampling to enhance sharp edges and variational token length to improve intricate geometric details. Hunyuan3D-DiT inherits the recent advanced flow matching models~\cite{lipman2022flow,esser2024scaling} to construct a scalable and flexible diffusion model.

For texture synthesis, Hunyuan3D-Paint introduces a multi-view PBR diffusion that generates albedo, metallic, and roughness maps for meshes. Notably, Hunyuan3D-Paint incorporates a spatial-aligned multi-attention module to align albedo and MR maps, 3D-aware RoPE to enhance cross-view consistency, and an illumination-invariant training strategy to produce light-free albedo maps robust to varying lighting conditions.

\emph{\methodName} separates shape and texture generation into distinct stages, an more advanced strategy proven effective upon previous large reconstruction models ~\cite{hong2023lrm,yang2024hunyuan3d,xu2024instantmesh,yang2024viewfusion,weng2023consistent123,liu2023syncdreamer,shi2023zero123++,liu2023zero}. This modularity allows users to generate untextured meshes only or apply textures to custom assets, enhancing flexibility for industrial applications.  

We rigorously evaluate \emph{\methodName} against leading commercial and recent open-source models, ~\eg, Michelangelo~\cite{zhao2023michelangelo}, Craftsman 1.5~\cite{li2024craftsman}, Trellis~\cite{xiang2024structured}, TripoSG~\cite{li2025triposg}, Step1X-3D~\cite{li2025step1x} and Direct3D-S2~\cite{wu2025direct3ds2gigascale3dgeneration}. Quantitative metrics and visual comparisons confirm its superiority in geometric detail preservation, texture-photo consistency, and human preference.  

This tutorial unpacks the architecture, data processing, training, and evaluation of \emph{\methodName}, providing practitioners with the tools to harness its capabilities for diverse 3D generation tasks.

\section{Data Processing}
In this section, we aim to describe the data processing for training the shape generation model and texture model. We start to introduce the dataset preparation, and then present how to obtain the relevant training and testing data for the shape generation model and texture model.

\subsection{Dataset collection}
For shape generation, we collect 100K+ textured and untextured 3D data from public datasets and custom datasets. The public dataset comes mainly from ShapeNet~\cite{chang2015shapenet}, ModelNet40~\cite{wu20153d}, Thingi10K~\cite{zhou2016thingi10k}, and Objaverse~\cite{deitke2023objaverse,deitke2023objaversexl}. For texture synthesis, we filter 70K+ human-annotated high-quality data following strict curation protocols from Objaverse-XL~\cite{deitke2023objaversexl}.

\subsection{Data preprocessing for shape generation}

\subsubsection{Normalization}
The normalization process begins by calculating the axis-aligned bounding box for each 3D object, ensuring all subsequent operations work in a standardized coordinate space. We apply uniform scaling to fit the object within a unit cube centered at the origin, preserving aspect ratios while maintaining consistent scale across the entire dataset. This spatial normalization is particularly crucial for neural networks to learn consistent geometric patterns, as it eliminates size variations that could otherwise dominate the learned features. For point cloud data, the implementation involves centering the cloud by subtracting its centroid, then scaling all points by the maximum Euclidean distance from the center, as shown in the provided Python snippet. This approach guarantees that all objects occupy approximately the same volume in the normalized space while preserving their original geometric relationships.

\subsubsection{Watertight}
The \textit{IGL} library generates watertight surfaces by constructing a signed distance field (SDF) from defective geometry. We initialize a uniform 3D query grid encompassing the input mesh. For each query point $\mathbf{q} \in Q_g$, IGL computes:  
$$\text{SDF}(\mathbf{q}) = \underbrace{\text{distance\_to\_mesh}(\mathbf{q}, V, F)}_{\text{nearest surface distance}} \cdot \underbrace{\text{sign}(\omega(\mathbf{q}))}_{\text{inside/outside sign}}$$
where $V$ and $F$ represent input vertices and faces. The sign is determined by the generalized winding number $\omega(\mathbf{q})$ where $\omega \approx 1$ indicates interior points and $\omega \approx 0$ exterior points.  

Sign consistency is enforced using IGL's winding number calculation. This resolves ambiguous signs near self-intersections by thresholding $\omega > 0.5$ for interior classification. The watertight mesh is extracted at the zero-level isosurface via marching cubes. The output $(V_\text{iso}, F_\text{iso})$ forms a topologically closed surface without boundary discontinuities.

\subsubsection{SDF Sampling}
In our approach, the creation of signed distance fields (SDF) serves as the core mathematical framework for representing 3D shapes. To achieve this, we employ a strategy of randomly selecting query points in two distinct ways: either close to the surface of the shape or evenly distributed throughout the entire $[-1,1]^3$ space. We then compute the SDF values for these points using the IGL computing library. The SDF values obtained from points near the surface are crucial for capturing the intricate details of the shape's surface. This allows the model to accurately represent fine features and subtle variations in the geometry. The SDF values from uniformly sampled points provide the model with a broader understanding of the overall structure and form of the 3D shapes. This dual sampling approach ensures that the model gains a comprehensive understanding of both detailed and general aspects of the shapes.

\subsubsection{Surface Sampling} Our hybrid sampling strategy combines the strengths of both uniform and feature-aware approaches to capture complete geometric information. Uniform sampling guarantees even coverage across the surface, forming approximately $50\%$ of the final point set. The remaining $50\%$ of points are strategically placed near high-curvature features through importance sampling based on local surface derivatives. The sampling density automatically adapts to geometric complexity, increasing point concentration in regions with intricate details while maintaining sparser sampling in simpler areas. This balanced approach ensures that sharp edges, corners, and other defining features receive adequate representation without unnecessarily dense sampling of planar regions, optimizing both the quality and efficiency of the resulting point set.

\subsubsection{Condition Render}
To render condition images for shape generation training, we sample 150 cameras uniformly distributed on a sphere centered at the origin using the Hammersley sequence algorithm with a randomized offset $\delta \in [0,1)^2$. An augmented dataset is generated with randomized FoVs $\theta_{\text{aug}} \sim \mathcal{U}(10^\circ, 70^\circ)$. in the meanwhile camera's  radius is adjusted between $r_{\text{aug}} \in [1.51, 9.94]$ to ensure consistent object framing.  

\begin{algorithm}
\caption{3D Data Preprocessing Pipeline}\label{alg:preprocess}
\begin{algorithmic}[1]
\Require Raw 3D mesh $X = (V,F)$ (vertices and faces)

\State \textbf{1. Normalization:}
\State $V_{norm} \gets Normalize(V)$

\State \textbf{2. Watertight Processing:}
\State Initialize empty SDF grid $\mathcal{G}$
\State $SDF \gets \text{IGL}(\mathcal{G}, V_{norm}, F)$
\State ($V_{iso}, F_{iso}) \gets \text{MarchingCube}(SDF, \text{level}=0)$

\State \textbf{3. SDF Sampling:}
\State $P_{surface} \gets \text{sample\_surface}(V_{iso}, F_{iso}, N_{near})$ \Comment{$N_{near}=249,856$ total points}
\State $P_{near} \gets \text{sample\_near\_surface}(V_{iso}, F_{iso}, N_{uniform})$ \Comment{$N_{uniform}=249,856$ total points}
\State Query points $P_{query} \gets P_{near} \cup P_{uniform}$ 
\State $ SDF_{query} \gets igl.signed\_distance(P_{query}, V_{iso}, F_{iso})$

\State \textbf{4. Surface Sampling:}
\State $P_{random} \gets \text{RandomSample}(V_{iso}, F_{iso}, N)$
\State $P_{sharp} \gets \text{SharpSample}(V_{iso}, F_{iso}, N)$ \Comment{$N=124928$ total points}

\State \textbf{5. Hammersley Condition Rendering:}
\State Generate Hammersley sequence $H_{150}$ on unit sphere
\State Apply random offset $\delta \sim \mathcal{U}([0,1)^2)$ to $H_{150}$
\For{each camera position $\mathbf{c}_i \in H_{150}$}
    \State Sample FoV $\theta_i \sim \mathcal{U}(10^\circ,70^\circ)$
    \State Compute radius $r_i \sim \mathcal{U}(\theta_{min},\theta_{max})$
    \State $Img_i \gets \text{render\_image}(X, \mathbf{c}_i, r_i)$
\EndFor

\State \Return $P_{query}, SDF_{query}, P_{random}, P_{sharp}, \{Img_i\}_{i=1}^{150}$
\end{algorithmic}
\end{algorithm}

\subsection{Data preprocessing for texture synthesis}


The texture synthesis heavily relies on 3D assets with rich texture details. Our training dataset consists of 70k+ human-annotated high quality data following strict curation protocols, which is filtered from Objaverse~\cite{deitke2023objaverse} and Objaverse-XL \cite{deitke2023objaversexl}. For each 3D object, we rendered data from four elevation angles: $-20^{\circ}$, $0^{\circ}$, $20^{\circ}$, and a random angle. At each elevation angle, we select 24 views that are uniformly distributed across azimuth dimension, generating corresponding albedo, metallic, roughness maps, and HDR/Point-light images of 512 $\times$ 512 resolution. We probabilistically render reference images using:
(1) Randomly sampled viewpoints (elevation: [-30°, 70°])
(2) Stochastic illumination: point lights (p=0.3) or HDR maps (p=0.7).

\section{Training}

\subsection{Hunyuan3D-Shape}
\label{subsec:shape}

Shape generation serves as the cornerstone of 3D generation, playing a crucial role in determining the usability of a 3D asset. Drawing inspiration from the success of the latent diffusion model~\cite{rombach2022high,zhang20233dshape2vecset, zhao2023michelangelo,zhang2024clay} in shape generation, we have adopted the generative diffusion model as the architecture for our shape model. Our shape generation model is composed of two main components: (1) an autoencoder, Hunyuan3D-ShapeVAE (Sec.~\ref{subsubsec:vae}), which compresses the shape of a 3D asset, represented by a polygon mesh, into a sequence of continuous tokens within the latent space; and (2) a flow-based diffusion model, Hunyuan3D-DiT (Sec.~\ref{subsubsec:dit}), which is trained on the latent space of ShapeVAE to predict object token sequences from a user-provided image. These predicted tokens are then decoded back into a polygon mesh using the VAE decoder. The specifics of these models are detailed below.

\begin{figure}
    \centering
    \includegraphics[width=0.94\linewidth]{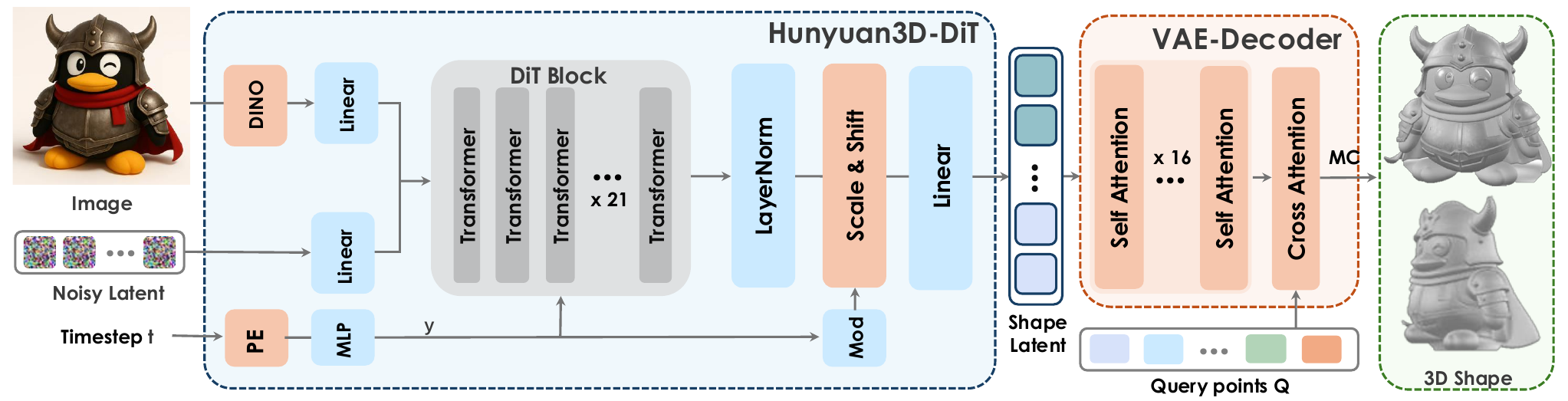}
    \caption{Overall pipeline for shape generation. Given a single image input, combining Hunyuan3D-DiT and Hunyuan3D-VAE can generate a high-quality and high-fidelity 3D shape.}
    \vspace{-4mm}
    \label{fig:dit}
\end{figure}

\begin{wrapfigure}{r}[1mm]{0pt}
\centering
\includegraphics[width=0.5\linewidth]{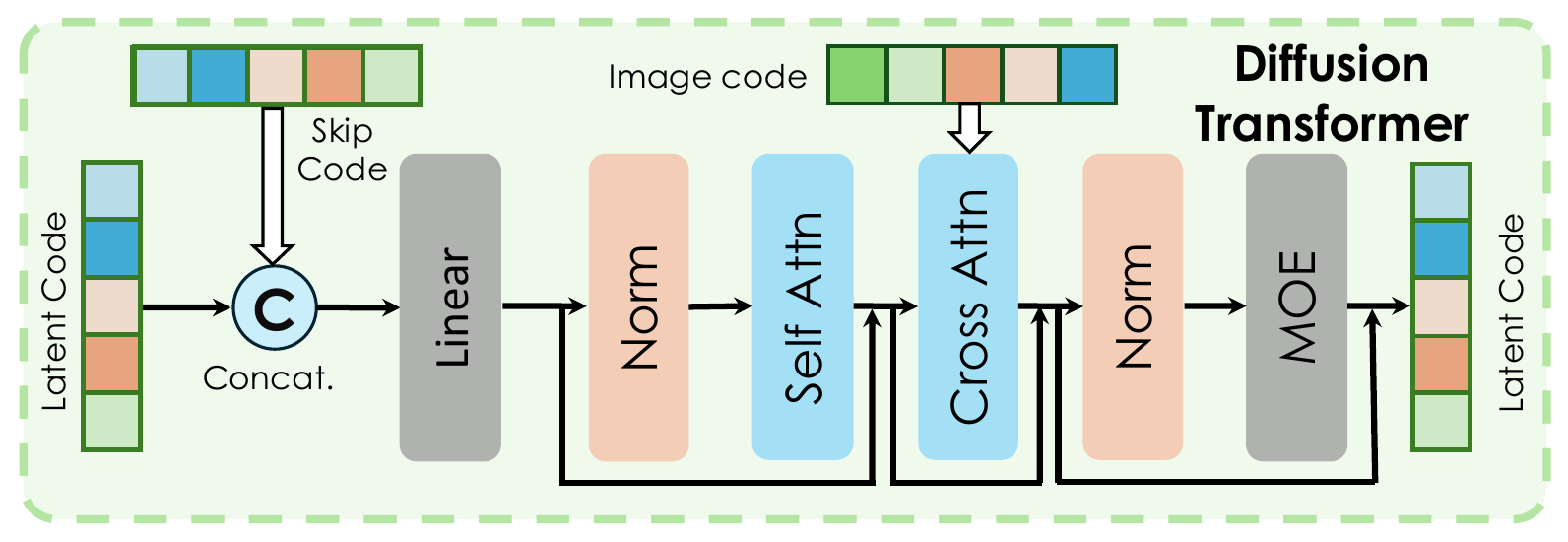}
\caption{Overview of DiT block. We adopt the DiT implemented by Hunyuan-DiT~\cite{li2024hunyuandit} in our pipeline.}
\label{fig:blocks}
\end{wrapfigure}

\subsubsection{Hunyuan3D-ShapeVAE}
\label{subsubsec:vae}

Hunyuan3D-ShapeVAE utilizes vector sets introduced by 3DShape2VecSet~\cite{zhang20233dshape2vecset}, and also used in the recent project Dora~\cite{zhang2024clay,chen2024dora}. Following these works, we employ a variational encoder-decoder transformer for compact shape representations. We use 3D coordinates and the normal vectors from point clouds sampled from the surfaces of 3D shapes as inputs for the encoder. The decoder is designed to predict the Signed Distance Function (SDF) of the 3D shape, which can be further transformed into a triangle mesh using the marching cube algorithm.


\textbf{Encoder.} For an input mesh, we first gather uniformly sampled surface point clouds $P_u \in \mathbb{R}^{M \times 3}$ and importance sampled point clouds $P_i \in \mathbb{R}^{N \times 3}$. The encoding process begins by applying Farthest Point Sampling (FPS) separately to $P_u$ and $P_i$ to generate query points $Q_u \in \mathbb{R}^{M' \times 3}$ and $Q_i \in \mathbb{R}^{N' \times 3}$ respectively. We then concatenate these points to form the final point cloud $P \in \mathbb{R}^{(M+N) \times 3}$ and query set $Q \in \mathbb{R}^{(M'+N') \times 3}$. Both $P$ and $Q$ are encoded by Fourier positional followed by linear projection, yielding encoded features $X_p \in \mathbb{R}^{(M+N) \times d}$ and $X_q \in \mathbb{R}^{(M'+N') \times d}$, where $d$ is the dimension. These features are processed through cross-attention and self-attention layers to obtain the hidden shape representation $H_s \in \mathbb{R}^{(M'+N') \times d}$. Following the variational autoencoder framework, we apply final linear projections to $H_s$ to predict the mean $\mathrm{E}(Z_s) \in \mathbb{R}^{(M'+N') \times d_0}$ and variance $\mathrm{Var}(Z_s) \in \mathbb{R}^{(M'+N') \times d_0}$ of the latent shape embedding, with $d_0$ being the latent dimension.

\textbf{Decoder.} The decoder $\mathcal{D}_s$ reconstructs a 3D neural field from the latent shape embedding $Z_s$. Initially, a projection layer maps the $d_0$-dimensional latent embedding to the transformer's hidden dimension $d$. Subsequent self-attention layers refine these embeddings, followed by a point perceiver module that queries a 3D grid $Q_g \in \mathbb{R}^{ (H \times W \times D) \times 3}$ to generate a neural field $F_g \in \mathbb{R}^{ (F_n \times W \times D) \times d}$. A final linear projection converts $F_g$ into a Sign Distance Function (SDF) $F_{sdf} \in \mathbb{R}^{ (F_o \times W \times D) \times 1}$, which is subsequently converted to a triangle mesh via marching cubes during inference.

\textbf{Training Strategy \& Implementation.}  We employ two losses to supervise the model training, including (1) the reconstruction loss that computes MSE loss between predicted SDF $\mathcal{D}_s (x | Z_s)$ and ground truth $\mathrm{SDF}(x)$, and (2) the KL-divergence loss $\mathcal{L}_{KL}$ to make the latent space compact and continuous. The overall training loss $\mathcal{L}_r$ can be written as,
\begin{equation}
    \mathcal{L}_r = \mathbb{E}_{ x \in \mathbb{R}^3 } [ \mathrm{MSE} ( \mathcal{D}_s (x | Z_s), \mathrm{SDF}(x) ) ] + \gamma \mathcal{L}_{KL}
\end{equation}
where $\gamma$ is the loss weight of KL loss.
To optimize computational efficiency, we implement a multi-resolution training strategy where latent token sequence lengths vary dynamically, with a maximum sequence length of 3072.

\subsubsection{Hunyuan3D-DiT}
\label{subsubsec:dit}
Hunyuan3D-DiT is a flow-based diffusion model designed to generate detailed and high-resolution 3D shapes based on image conditions.

\textbf{Condition encoder.} To capture detailed image features, we employ a large image encoder, DINOv2 Giant~\cite{oquab2023dinov2} with an image size of $518 \times 518$. Additionally, we remove the background from the input image, resize the object to a standard size, center it, and fill the background with white.

\textbf{DiT block.} Inspired by Hunyuan-DiT~\cite{li2024hunyuandit} and TripoSG~\cite{li2025triposg}, we adopt transformers structure as shown in Fig.~\ref{fig:dit}. We stack the 21 Transformer layers to learn the latent codes. As shown in Fig.~\ref{fig:blocks}, in each Transformer layer, we leverage the dimension concatenation to introduce the skip connection of the latent code. Similar to previous methods~\cite{zhang2024clay, li2024craftsman}, we employ the cross-attention layer to project the image condition into the latent code. In addition, an MOE layer is used to enhance the representation learning of the latent code.

\textbf{Training \& Inference.} We train our model using the flow matching objective~\cite{lipman2022flow,esser2024scaling}. Flow matching defines a probability path between Gaussian and data distributions, training the model to predict the velocity field $u_t = \frac{x_t}{d_t}$ that moves sample $x_t$ towards data $x_1$. We use the affine path with a conditional optimal transport schedule as specified in~\cite{lipman2024flowmatchingguidecode}, where $x_t = ( 1-t ) \times x_0 + t \times x_1$, $u_t = x_1 - x_0$. The training loss is formulated as,
\begin{equation}
    \mathcal{L} = \mathbb{E}_{t,x_0,x_1} [ \parallel u_\theta(x_t,c,t) - u_t \parallel_2^2 ],
\end{equation}
where $t \sim \mathbb{U}(0,1)$ and $c$ represents model condition. During inference, we randomly sample a starting point and use a first-order Euler ordinary differential equation (ODE) solver to compute $x_1$ with our diffusion model $u_\theta(x_t,c,t)$.

\subsection{Hunyuan3D-Paint}
Traditional color textures are no longer sufficient to meet the demands for photorealistic 3D asset generation. Therefore, we introduce a PBR material texture synthesis framework advancing beyond conventional RGB texture maps. We adhere to the BRDF model and simultaneously output albedo, roughness, and metallic maps from multiple viewpoints, aiming to accurately describe the surface reflectance properties of generated 3D assets and precisely simulate the distribution of geometric micro-surfaces, resulting in more realistic and detailed rendering effects. Further, we introduce 3D-Aware RoPE to inject spatial information, significantly improving cross-view consistency and enabling seamless texturing.

\begin{figure}[h]
    \centering
    \includegraphics[width=1\textwidth]{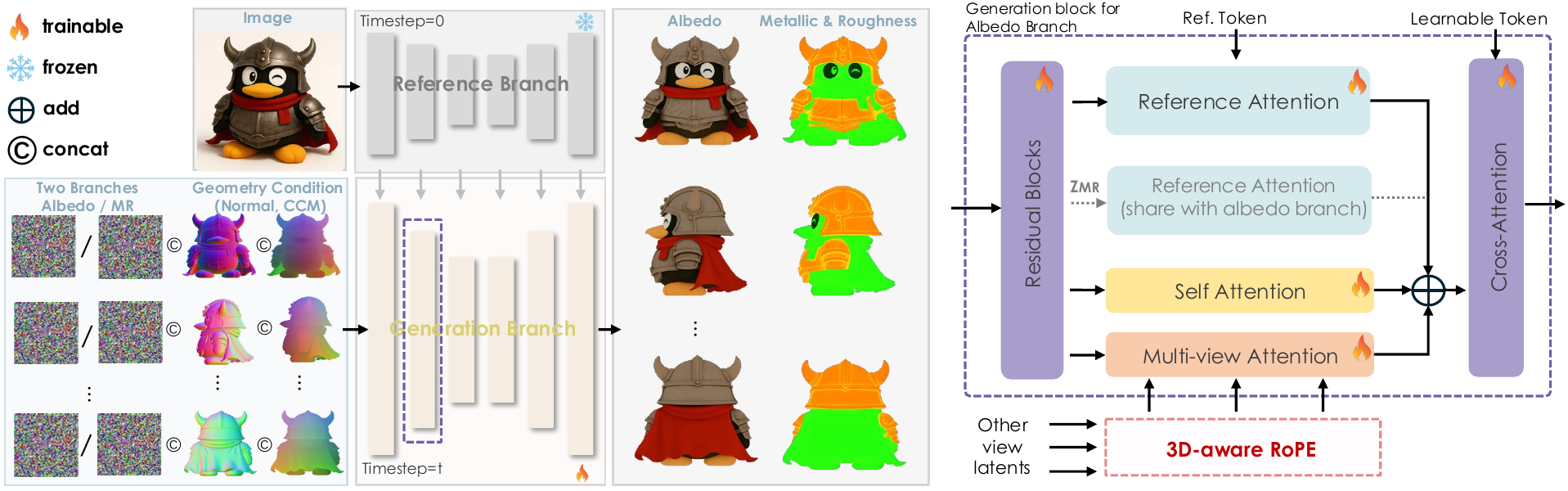}
    \caption{Overview of material generation framework. }
    \label{fig:texture_pipe}
\end{figure}

\textbf{Basic Architecture.} Building on the multiview texture generation architecture of Hunyuan3D-2~\cite{zhao2025hunyuan3d}, we introduce a novel material generation framework, as is shown in the left side of Fig.\ref{fig:texture_pipe}. The framework implements the Disney Principled BRDF model~\cite{burley2012physically} to generate high-quality PBR material maps.
We retain the reference image feature injection mechanism of ReferenceNet, while concatenating both geometry-rendered normal maps and CCM (canonical coordinate map) with latent noise. 

\textbf{Spatial-Aligned Multi-Attention Module.} We employ a pre-trained VAE for multi-channel material image compression while implementing a parallel dual-branch UNet architecture~\cite{he2025materialmvpilluminationinvariantmaterialgeneration} for material generation. As shown in the right side of Fig.\ref{fig:texture_pipe}, we implement parallel multi-attention modules~\cite{feng2025romantexdecoupling3dawarerotary} comprising self-attention, multi-view attention, and reference attention for both albedo and metallic-roughness (MR) maps. To model the physical relationships between albedo/MR maps and reference images, and to achieve spatial alignment between MR and albedo maps, we directly propagate the computed outputs from the albedo reference attention module to the MR branch.

\textbf{3D-Aware RoPE.} To address texture seams and ghosting artifacts caused by local inconsistencies across neibor views, 3D-Aware RoPE~\cite{feng2025romantexdecoupling3dawarerotary} is introduced into the multiview attention block for enhanced cross-view coherence. Specifically, by downsampling the 3D coordinate volume, we construct multi-resolution 3D coordinate encodings aligned with the UNet hierarchy levels. These encodings are additively fused with corresponding hidden states, thereby integrating cross-view interactions into 3D space to enforce multi-view consistency.

\textbf{Illumination-Invariant Training Strategy.}
To generate light- and shadow-free albedo map and accurate MR map, we posit an intuitive insight: while rendering results of the same object differ under diverse lighting, its intrinsic material properties should remain consistent. Consequently, we design an illumination-invariant training strategy~\cite{he2025materialmvpilluminationinvariantmaterialgeneration} to enforce this property. Specifically, consistency loss is computed by adopting two sets of training samples containing reference images of the same object rendered by different lighting conditions.

\textbf{Experimental Setup.} Our model is initialized from the Zero-SNR checkpoint~\cite{lin2024common} of Stable Diffusion 2.1 and optimized using the AdamW at a learning rate of $5 \times 10^{-5}$. The training protocol incorporates 2000 warm-up steps, requiring approximately 180 GPU-days.

\section{Evaluation}
To assess the effectiveness of a 3D generative model, we conduct experiments focusing on three key areas: (1) 3D Shape Generation (untextured shape evaluation), (2) Texture Synthesis, and (3) Complete 3D asset creation (textured 3D shapes).

\begin{table}
\centering
\small
\begin{tabular}{l|cccccc}
\hline
Models & ULIP-T ($\uparrow$) & ULIP-I ($\uparrow$) & Uni3D-T ($\uparrow$)& Uni3D-I ($\uparrow$) \\ \hline
Michelangelo~\cite{zhao2023michelangelo} & 0.0752 & 0.1152 & 0.2133 & 0.2611 \\
Craftsman 1.5~\cite{li2024craftsman}    & 0.0745 & 0.1296 & 0.2375 & 0.2987 \\
TripoSG~\cite{li2025triposg} & 0.0767 &  0.1225 &  0.2506 & 0.3129 \\
Step1X-3D~\cite{li2025step1x}    &   0.0735 & 0.1183 & 0.2554 & 0.3195 \\
Trellis~\cite{xiang2024structured} & 0.0769 & 0.1267 & 0.2496 & 0.3116 \\
Direct3D-S2~\cite{wu2025direct3ds2gigascale3dgeneration} & 0.0706 & 0.1134 & 0.2346 & 0.2930\\
Hunyuan3D-DiT & \textbf{0.0774} & \bf{0.1395} & \textbf{0.2556} & \textbf{0.3213} \\ \hline
\end{tabular}
\vspace{3mm}
\caption{The quantitative comparison for shape generation. The Hunyuan3D-DiT presents the best performance.}
\label{tab:shapegen_transposed}
\end{table}

\begin{figure}[t]
    \centering
    \includegraphics[width=\linewidth]{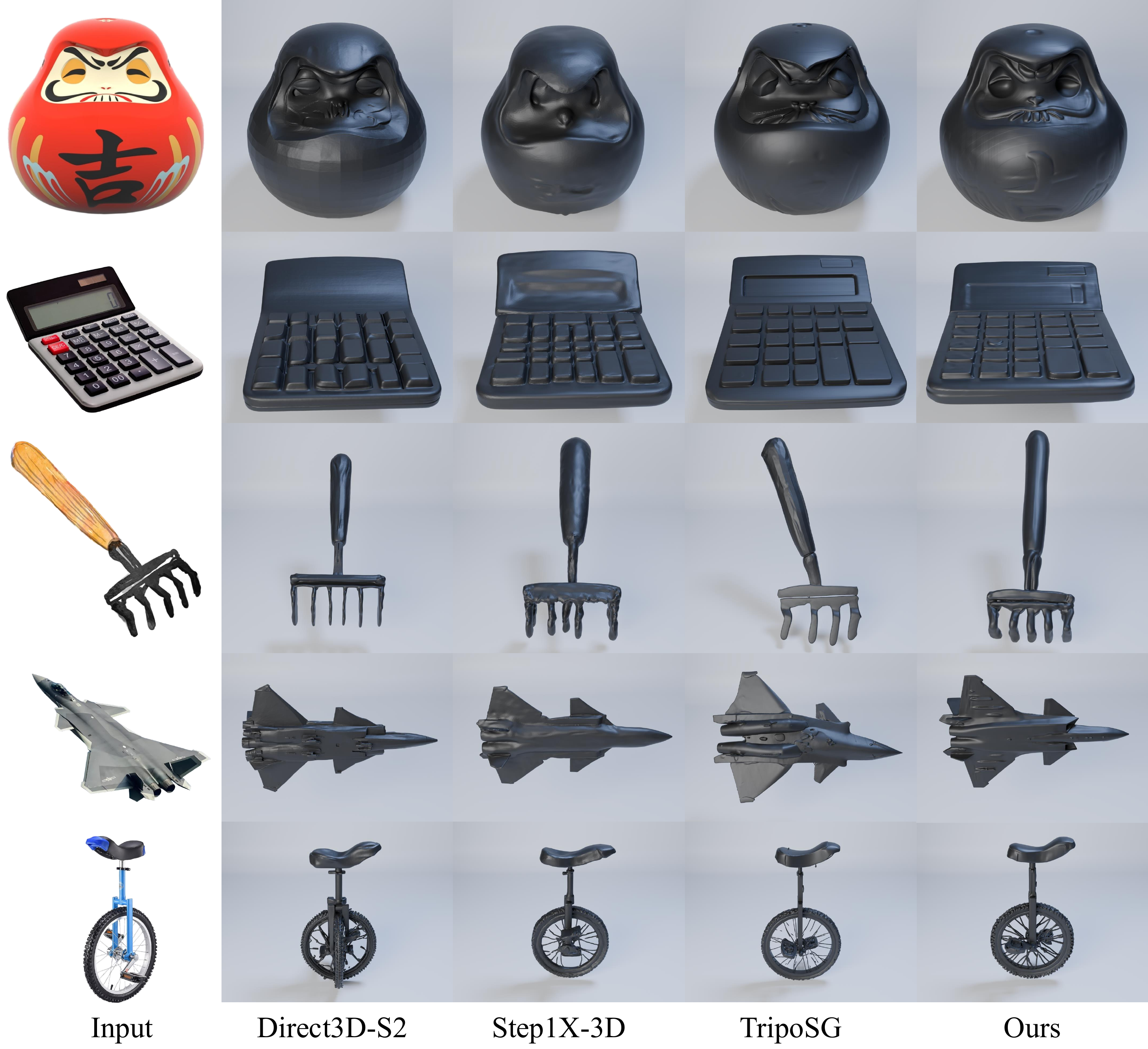}
    \caption{The qualitative comparisons for image-to-shape generation.}
    \label{fig:geo_compare}
\end{figure}

\subsection{3D Shape Generation}
Shape generation is crucial for 3D generation, as detailed and high-resolution meshes provide the groundwork for subsequent tasks. In this section, we evaluate the 3D shape generation capabilities of Hunyuan3D-DiT, focusing on shape creation.

\textbf{Metrics.} To evaluate shape generation performance, we used ULIP~\cite{xue2023ulip} and Uni3D~\cite{zhouuni3d} to measure the similarity between the generated mesh and input images. Specifically, we sampled 8,192 surface points from the generated mesh as the point cloud modality. We then utilized the caption of the input image obtained from an existing VLM model as the text modality. Finally, we applied the ULIP models to obtain the ULIP-I and ULIP-T scores, which measure the similarity between the point cloud and text, as well as the similarity between the point cloud and image, respectively. In this context, the text caption comes from a VLM model. We also employed the same process to obtain the Uni3D-I and Uni3D-T scores based on the Uni3D model.

\textbf{Comparison with Shape Generation Models.} We compared Shape Quality with several leading models, including open-source options like Direct3D-S2~\cite{wu2025direct3ds2gigascale3dgeneration}, Step1X-3D~\cite{li2025step1x}, and TripoSG~\cite{li2025triposg}. ~\Cref{tab:shapegen_transposed} presents a numerical comparison between Hunyuan3D-DiT and other methods, showing that Hunyuan3D-DiT delivers the most accurate results. Additionally, the visual comparison in Fig.~\ref{fig:geo_compare} confirms the adherence of Hunyuan3D-DiT to image prompts, including the faithful capture of intricate details (details of roly-poly toys, the number of calculator buttons, the number of teeth on a rake, and the structure of a fighter jet), and its ability to produce watertight meshes ready for downstream applications.

\subsection{Texture Map Synthesis}
\label{sec:texture_sythesis}

As texture maps directly influence the visual appeal of textured 3D assets, we conduct comprehensive quantitative and qualitative comparisons of texture generation methodologies across both academic and industrial domains.

\begin{figure}[h]
    \centering
    \includegraphics[width=0.8\linewidth]{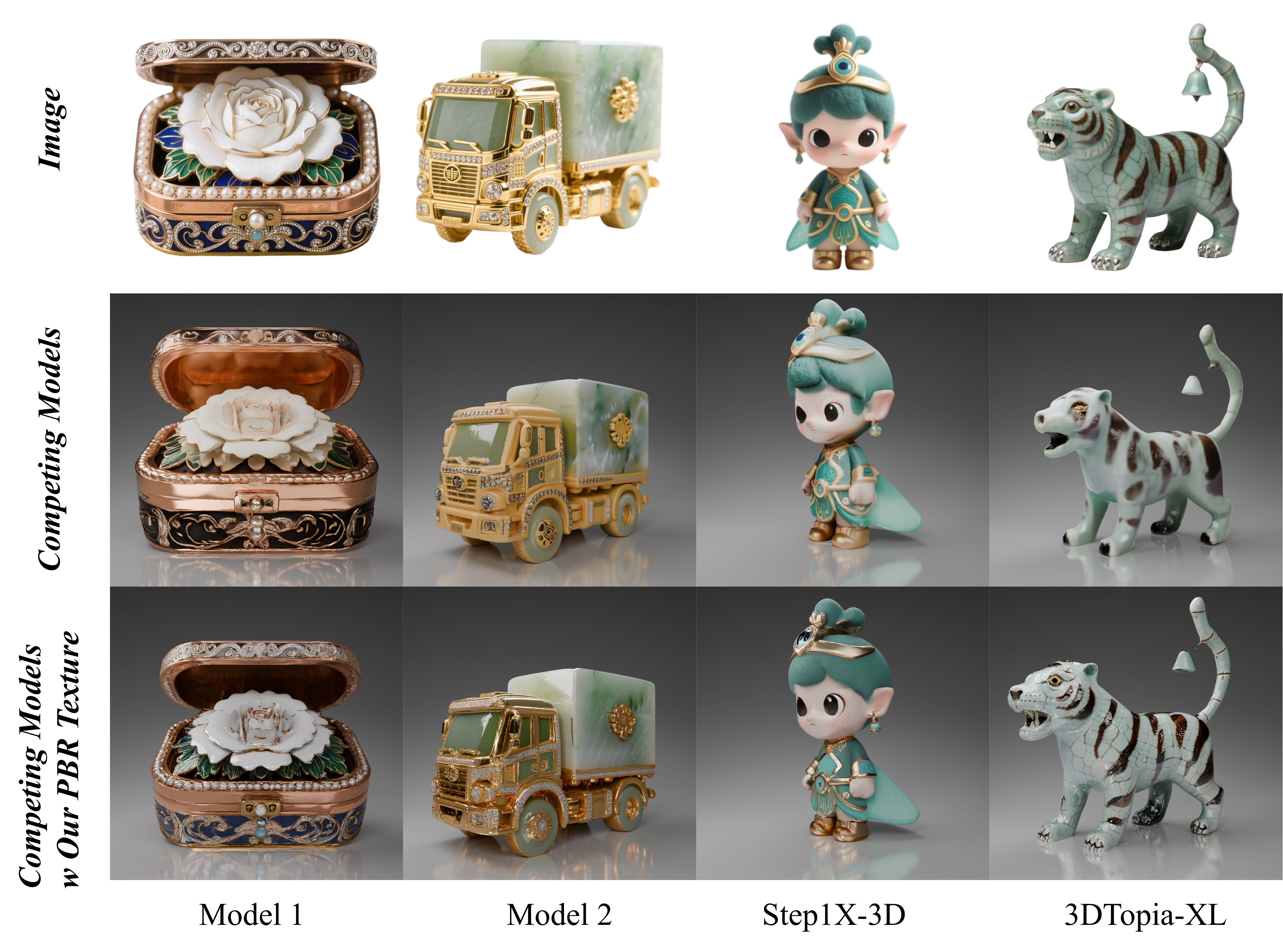}
    \caption{The qualitative comparisons for texture synthesis.}
    \label{fig:tex_comp}
\end{figure}

\textbf{Comparison with Texture Synthesis Models.} To quantify the similarity between generated textures and ground truth, we employ Fréchet Inception Distance (FID)~\cite{heusel2017gans}, CLIP-based FID (CLIP-FID)~\cite{radford2021learning}, and Learned Perceptual Image Patch Similarity (LPIPS)~\cite{zhang2018unreasonable} metrics on Hunyuan3D-Paint and baseline image-to-texture models, including SyncMVD-IPA~\cite{liu2024text}, TexGen~\cite{yu2024texgen} and Hunyuan3D-2.0~\cite{zhao2025hunyuan3d}. Given an untextured shape and a single image, we compare these models with our results through both quantitative and qualitative evaluations. The quantitative results are shown in Tab.~\ref{tab:img2tex}, while the qualitative results are illustrated in Figure~\ref{fig:tex_comp}. These evaluations clearly demonstrate the superiority of our method over all comparative approaches.

\begin{table}
\centering
\small
\begin{tabular}{l|cccccc}
\toprule
\textbf{Method} & {CLIP-FID ($\downarrow$}) & {CMMD ($\downarrow$)} & {CLIP-I ($\uparrow$)} & {LPIPS ($\downarrow$)} \\
\midrule
SyncMVD-IPA~\cite{liu2024text}     & 28.39 & 2.397 & 0.8823 & 0.1423 \\
TexGen~\cite{yu2024texgen}         & 28.24 & 2.448 & 0.8818 & 0.1331 \\
Hunyuan3D-2.0~\cite{zhao2025hunyuan3d} & 26.44 & 2.318 & 0.8893 & 0.1261 \\
Hunyuan3D-Paint & \bf{24.78} & \bf{2.191} & \bf{0.9207} & \bf{0.1211} \\
\bottomrule
\end{tabular}
\caption{The quantitative comparison for texture generation. Hunyuan3D-Paint achieves the best performance.}
\label{tab:img2tex}
\end{table}

\begin{figure}[h]
    \centering
    \includegraphics[width=\linewidth]{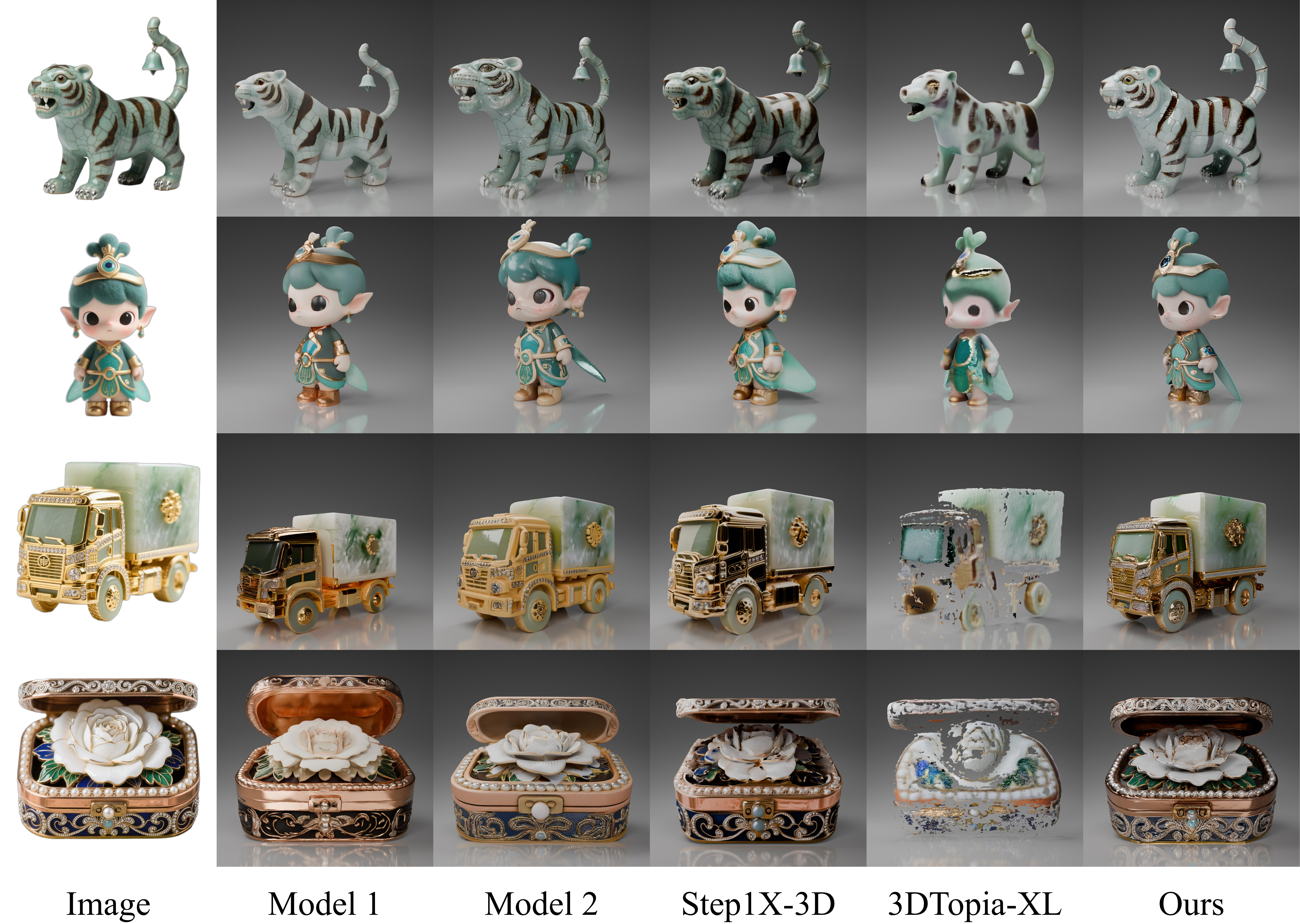}
    \vspace{-5mm}
    \caption{The qualitative comparisons for image-to-3D generation.}
    \label{fig:e2e}
\end{figure}

\textbf{Comparison with Image-to-3D Models.}
We also conduct visualized comparison with publicly accessible 3D generation algorithms, including the open-source Step1X-3D~\cite{li2025step1x} and 3DTopia-XL~\cite{chen20243dtopia}, alongside the commercial Model 1 and Model 2. Given a single image, all compared methods can form geometry and corresponding PBR material maps. Specifically, we assess the end-to-end quality across these methods, as shown in Fig.~\ref{fig:e2e}. These results demonstrate that our model not only generates PBR material maps with the highest fidelity but also effectively mitigates shortcomings associated with lower-quality geometries. This leads to superior end-to-end performance compared to existing methods.

\section{Conclusion}
\emph{\methodName} introduces a groundbreaking approach for production-ready 3D content creation by unifying high-fidelity geometry generation and PBR material synthesis within an open-source framework. Its architecture, which combines a DiT for shape generation and a multi-view conditioned painter for PBR material synthesis, allows for the rapid creation of studio-quality assets with exceptional visual fidelity. By open-sourcing the entire data processing, training pipelines, and model weights, this system makes advanced 3D AIGC accessible to a wider audience, revolutionizing workflows in gaming, virtual reality, and industrial design. Quantitative metrics demonstrate its superiority in both geometric accuracy and material quality. As the first fully open-sourced solution for PBR-textured 3D asset generation, \emph{\methodName} bridges the gap between academic research and scalable content creation, encouraging global collaboration to shape the future of 3D generative AI.

\clearpage
\section{Contributors}
\begin{itemize}[leftmargin=0.25cm]
  \item \textbf{Project Sponsors:} Jie Jiang, Linus, Yuhong Liu, Di Wang, Tian Liu, Peng Chen
    \item \textbf{Project Leaders:} Chunchao Guo, Jingwei Huang
    \item \textbf{Core Contributors:}
    \begin{itemize}[leftmargin=0.5cm]
        \item \textbf{Data:} Lifu Wang, Sicong Liu, Jihong Zhang, Meng Chen, Liang Dong, Yiwen Jia, Yulin Cai, Jiaao Yu, Yixuan Tang, Dongyuan Guo, Junlin Yu, Hao Zhang, Zheng Ye, Peng He, Runzhou Wu, Shida Wei, Chao Zhang, Yonghao Tan
        \item \textbf{Shape Generation:} Haolin Liu, Yunfei Zhao, Qingxiang Lin, Zeqiang Lai, Xianghui Yang, Huiwen Shi, Zibo Zhao, Bowen Zhang, Hongyu Yan
        \item \textbf{Texture Synthesis:} Shuhui Yang, Mingxin Yang, Yifei Feng, Xin Huang, Sheng Zhang, Zebin He, Di Luo
        \item \textbf{Infra: } Yifu Sun, Lin Niu, Shirui Huang, Bojian Zheng, Shu Liu, Shilin Chen, Xiang Yuan, Xiaofeng Yang, Kai Liu, Jianchen Zhu
    \end{itemize}
\end{itemize}
\clearpage
{\small
\bibliographystyle{unsrt}
\bibliography{neurips_2025}
}
\clearpage


\end{document}